\documentclass[conference]{IEEEtran}

\usepackage{graphicx}
\usepackage{amsmath,amssymb}
\usepackage{bm}
\usepackage{stfloats}
\usepackage{cuted}
\usepackage{capt-of}
\usepackage{hyperref}

\makeatletter
\let\@oldmaketitle\@maketitle
\renewcommand{\@maketitle}{\@oldmaketitle
   \begin{center}
      \setcounter{figure}{0}
      \includegraphics[width=0.95\linewidth]{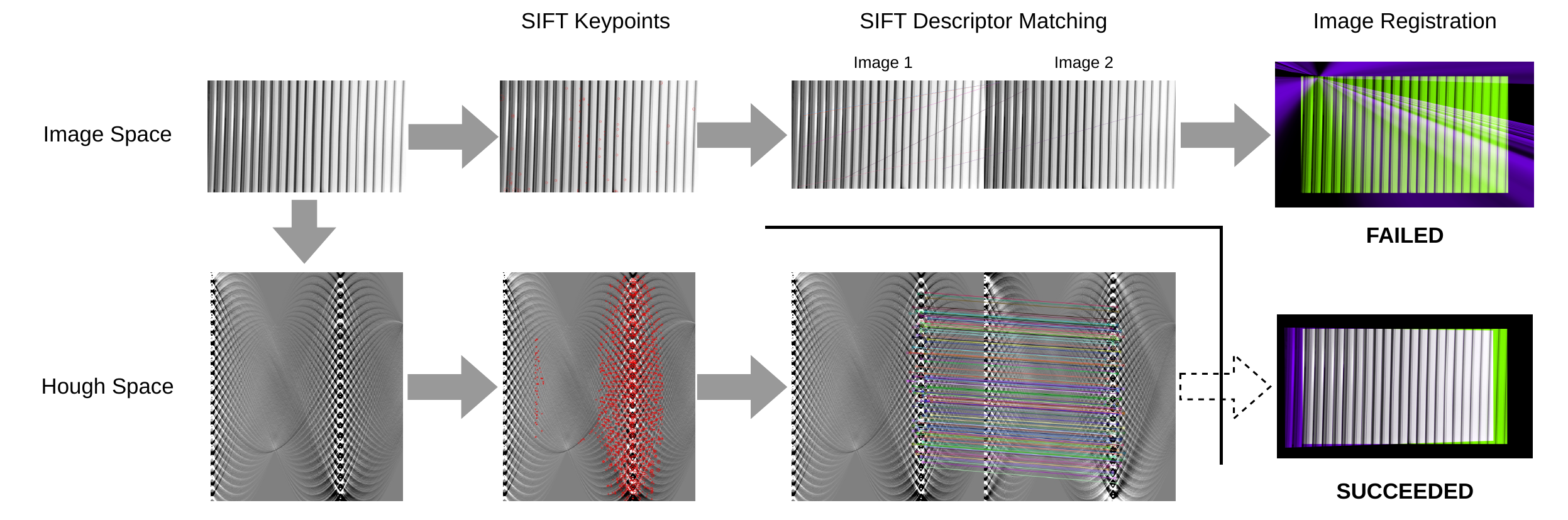}
      \captionof{figure}{Overview of the proposed Hough-SIFT method. First, the input image is transformed into Hough space using the proposed texture-preserving Hough transform. Next, SIFT keypoints are extracted from the Hough image, and descriptor matching is performed. Finally, homography is estimated from the correspondences in Hough space. Only RANSAC inliers are shown in the descriptor matching results for clarity. In the image registration column, image 1 (green) is transformed to align with image 2 (purple) based on the estimated homography. The monochrome areas in the result indicate successful alignment. While conventional SIFT fails, Hough-SIFT successfully registers the images.}
      \label{fig:overview}
  \end{center}
}
\makeatother

\DeclareMathOperator*{\argmin}{arg\,min}

\begin{document}
\pagestyle{plain}
\thispagestyle{plain}

\title{Hough-SIFT: Robust Image Registration \\ for Linear Structures via Hough Space}
\author{\IEEEauthorblockN{Masaki Satoh}
\IEEEauthorblockA{
\textit{Morpho, Inc.}\\
Tokyo, Japan \\
m-satoh@morphoinc.com}
}

\maketitle

\IEEEpeerreviewmaketitle

\begin{abstract}
Image registration is essential in applications such as electronic image stabilization. Scale-Invariant Feature Transform (SIFT), a widely used local keypoint detector and descriptor, typically provides accurate registration; however, it often fails in scenes with strong linear structures (e.g., shutters), where local features become ambiguous. We propose Hough-SIFT, a robust registration method that performs SIFT descriptor matching in Hough space. In this domain, linear structures form distinctive peaks that restore descriptor discriminability. Experiments demonstrate that Hough-SIFT is robust in linear scenes where SIFT frequently fails, while maintaining accuracy comparable to SIFT in normal scenes.
\end{abstract}

\section{Introduction}
Many computer vision applications require image registration to estimate geometric transformations between images. An important example is electronic image stabilization (EIS)\cite{morikawa2021}, wherein consecutive frames are aligned to suppress camera shake. Scale-Invariant Feature Transform (SIFT)\cite{lowe2004sift} is among the most widely used local feature detectors and descriptors for this task. However, SIFT often fails when scenes contain strong linear structures, such as louvers or shutters. In such scenes, many similar local patterns occur; consequently, SIFT detects few keypoints with ambiguous descriptors, leading to mismatches and registration failure.

Man-made linear structures are ubiquitous in everyday environments and frequently appear in captured images. In video processing, even a single registration failure can degrade the final result. Therefore, robust registration in such scenes is essential.

We propose Hough-SIFT to address this limitation. Rather than applying SIFT in the image domain, we apply it to a Hough-space representation\cite{duda1972hough}. Figure~\ref{fig:overview} summarizes the pipeline. The key insight is that linear structures form distinctive peaks in Hough space, thereby restoring descriptor discriminability. We introduce a texture-preserving Hough transform and a cost function for homography estimation in Hough space. Experiments on synthetic datasets and real-world videos demonstrate that Hough-SIFT outperforms conventional SIFT in challenging scenes while maintaining comparable accuracy in normal scenes.

Line segment detectors such as the Line Segment Detector\cite{von2010lsd} and line-feature matching methods such as the Line Band Descriptor\cite{zhang2013lbd} have been applied to registration in linear scenes. However, line matching often requires complex geometric verification or graph-based matching to handle fragmentation and occlusion. In contrast, Hough-SIFT leverages the well-established SIFT framework, enabling straightforward descriptor matching and homography estimation with standard techniques.

Learning-based registration methods have been recently proposed. For instance, GlueStick\cite{pautrat2023gluestick} combines point and line features for robust registration. However, such methods generally require large training datasets, may not generalize well to scenes dominated by linear structures (which are underrepresented in typical training data), and often prove computationally expensive for real-time applications. In contrast, Hough-SIFT requires no training and can be applied directly to arbitrary images.

\section{Proposed Method}
Hough-SIFT comprises three main steps. First, the input images are transformed into Hough space. Next, SIFT keypoints are extracted and matched in the Hough domain. Finally, homography is estimated from the resulting correspondences. This section details the first and third steps because the second step follows standard SIFT procedures.

\subsection{Texture-Preserving Hough Transform}
In scenes dominated by linear structures, SIFT-like keypoint detectors struggle because their discriminative power depends on local texture, which is often limited. To overcome this limitation, global structural information, such as lines, must be incorporated.

The Hough transform is a well-established technique for line detection\cite{duda1972hough}. A point in the Hough image corresponds to a line in the original image. In the standard Hough transform, edge points are detected using methods such as the Canny edge detector\cite{canny1986edge}, and pixel values in the Hough space correspond to the number of edge points on the corresponding line in the original image. In this way, lines are represented as extrema.

\begin{figure}[htb]
  \centering
  \includegraphics[width=0.35\linewidth, trim = 50mm 50mm 110mm 45mm, clip]{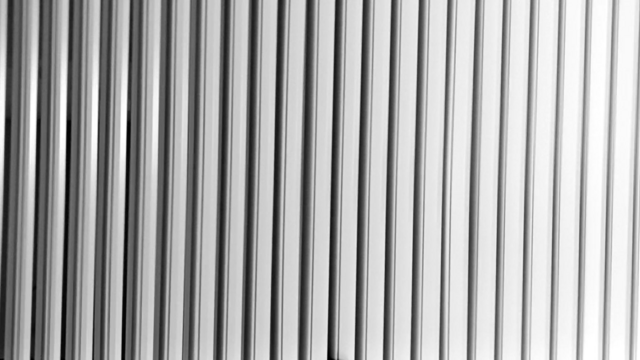}
  \includegraphics[width=0.35\linewidth, trim = 50mm 50mm 110mm 45mm, clip]{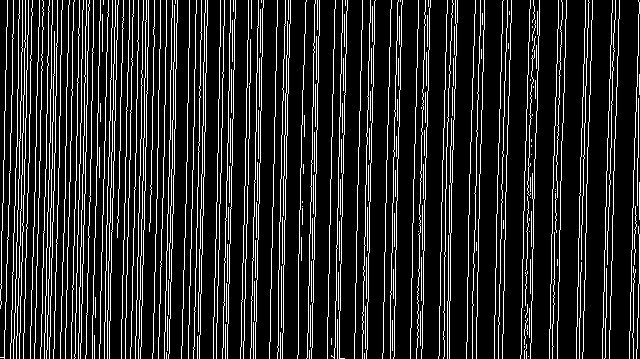}
  \caption{(Left) An example image of linear structures. (Right) The result of applying the Canny edge detector to the left image. In the original image, distinguishing individual linear structures is difficult but still possible thanks to subtle textures and patterns between the lines. However, after applying the Canny detector, the edge points become sparse and fail to capture these texture variations, making it difficult to distinguish different lines.}
  \label{fig:canny}
\end{figure}

The Hough transform acts as a texture-localization operator: it compresses information along lines in the image into points in Hough space. Consequently, linear structures are localized, allowing SIFT to recover discriminative power. However, the standard edge-counting Hough transform discards much of the texture information. \figurename~\ref{fig:canny} shows that distinguishing different lines becomes difficult after Canny edge detection. To address this issue, we propose a texture-preserving Hough transform that retains texture information while localizing linear structures.

Let $(x, y)$ denote coordinates in the original image $I(x, y)$, and let $(\theta, r)$ denote coordinates in the Hough image $J(\theta, r)$, where $\theta$ is the angle of the line normal vector $\bm{n}(\theta) \equiv (\cos\theta, \sin\theta)$ and $r$ is the distance from the origin to the line along $\bm{n}(\theta)$. A line $\bm{l}(\theta, r)$ in the image space can be represented as $x \cos\theta + y \sin\theta - r = 0$. We propose the following texture-preserving Hough transform:
\begin{align}
J(\theta,r) =
  \sum_{(x,y) \in \bm{l}(\theta,r)}
  \bm{n}(\theta) \cdot \nabla I(x,y).
\label{eq:hough_transform}
\end{align}
The directional derivative $\bm{n}(\theta)\cdot\nabla I(x,y)$ responds strongly to linear structures along $\bm{l}(\theta,r)$; consequently, lines appear as extrema in the Hough image, similar to the standard transform. Additionally, texture around these lines is preserved in pixel values near the extrema, enabling SIFT to extract more discriminative features. This transform is related to the Radon transform\cite{radon1917} and can be interpreted as the $r$-derivative thereof.

The parameter ranges in Hough space require careful selection. Although a naive choice would be $\theta \in [0, 2\pi)$ and $r \in [0, D/2]$, where $D$ is the image diagonal, it is computationally convenient to allow negative $r$ and restrict $\theta$ to $[0, \pi)$, since $\bm{l}(\theta,r)$ and $\bm{l}(\theta+\pi,-r)$ represent the same line. Notably, these configurations produce opposite signs in Hough space: $J(\theta + \pi, -r) = -J(\theta, r)$. To avoid discontinuities, we extend $\theta$ to $[0, 5\pi/4)$ and use keypoints in $[\pi/8, 9\pi/8)$ from one image and in $[0, 5\pi/4)$ from the other. Additionally, to reduce computational waste from minor boundary lines, we restrict $r$ to $(-L/2, L/2)$, where $L \equiv \max(\text{width}, \text{height})$.

\subsection{Homography Estimation in Hough Space}
In 2D projective geometry, lines are dual to points\cite{hartley2003multiple}. When a point $(x, y, 1)^\mathrm{T}$ in homogeneous coordinates is transformed by a homography matrix $H$, a line $(a, b, c)^\mathrm{T}$ is transformed by the inverse transpose of the homography matrix, $H^{-\mathrm{T}}$. Therefore, we can estimate the homography between two images from the correspondences in Hough space.

Let $(\theta_1^i,r_1^i)$ and $(\theta_2^i,r_2^i)$ denote the coordinates of the $i$-th matched keypoint pair in the Hough spaces of images 1 and 2, respectively. Let $(\tilde{\theta}_2^i(H),\tilde{r}_2^i(H))$ be the transformed coordinates of $(\theta_1^i,r_1^i)$ under homography $H$. We define the cost as the sum of squared distances in Hough space and estimate $H$ by minimizing
\begin{align}
  \hat{H} = \argmin_H \sum_i \bm{\epsilon}^i(H)^\mathrm{T} \bm{\epsilon}^i(H),
\end{align}
where $\bm{\epsilon}^i(H)$ denotes the residual vector for the $i$-th correspondence, defined as
\begin{align}
  \bm{\epsilon}^i(H) \equiv
  \begin{bmatrix}
  \Delta \theta^i(H) \\
  \Delta r^i(H)
  \end{bmatrix} \equiv
  \begin{bmatrix}
    \tilde{\theta}_2^i(H) - \theta_2^i \\
    \tilde{r}_2^i(H) - r_2^i
  \end{bmatrix}.
  \label{eq:residual0}
\end{align}
A limitation of this definition is that the periodicity of angle $\theta$ must be handled explicitly.

Instead of using the Hough space coordinates directly, we convert them back to homogeneous coordinates in the original image: $\bm{l}(\theta,r) =(\cos\theta, \sin\theta, -r)^\mathrm{T}$. Then, we define the residual vector as:
\begin{align}
  \bm{\epsilon}^i(H) \equiv
  \tilde{\bm{l}}_2^i(H) - \bm{l}_2^i,
  \label{eq:residual1}
\end{align}
where $\tilde{\bm{l}}_2^i(H)\equiv (\cos\tilde{\theta}_2^i(H), \sin\tilde{\theta}_2^i(H), -\tilde{r}_2^i(H))^\mathrm{T}$ is the transformed line in homogeneous form. This formulation avoids periodicity issues. For inliers, the residual norm is expected to be small; assuming $|\Delta \theta^i(H)|\ll 1$ and $|\Delta r^i(H)|\ll 1$ and applying first-order Taylor expansion yields
\begin{align}
  \bm{\epsilon}^i(H) \simeq
  \begin{bmatrix}
    -\sin\theta_2^i \, \Delta \theta^i(H) \\
    \cos\theta_2^i \, \Delta \theta^i(H) \\
    - \Delta r^i(H)
  \end{bmatrix},
\end{align}
with norm $\|\bm{\epsilon}^i(H)\|^2 \simeq (\Delta \theta^i(H))^2 + (\Delta r^i(H))^2$. Thus, Eq.~(\ref{eq:residual1}) approximately reproduces Eq.~(\ref{eq:residual0}) while avoiding periodicity issues, and we adopt the latter throughout.

With this cost function, Hough-SIFT estimates homography using standard nonlinear optimization: inlier selection with Random Sample Consensus (RANSAC)\cite{fischler1981ransac}, followed by Levenberg--Marquardt refinement.

\section{Experiments and Discussion}

We evaluated Hough-SIFT on synthetic image pairs using geometric and photometric errors, and on real-world videos using photometric error. We compared it with conventional SIFT.

We used the OpenCV implementations of SIFT with default settings and converted input images to Hough images with resolution $(480+120)\times640$, where $480$ corresponds to the $\theta$ axis and $640$ to the $r$ axis, and $+120$ is padding on the $\theta$ axis to avoid boundary effects.

\subsection{Synthetic Image Pairs}

\begin{figure}[!htb]
  \centering
  \begin{tabular}{cc}
    \begin{minipage}{0.45\linewidth}
      \includegraphics[width=\linewidth]{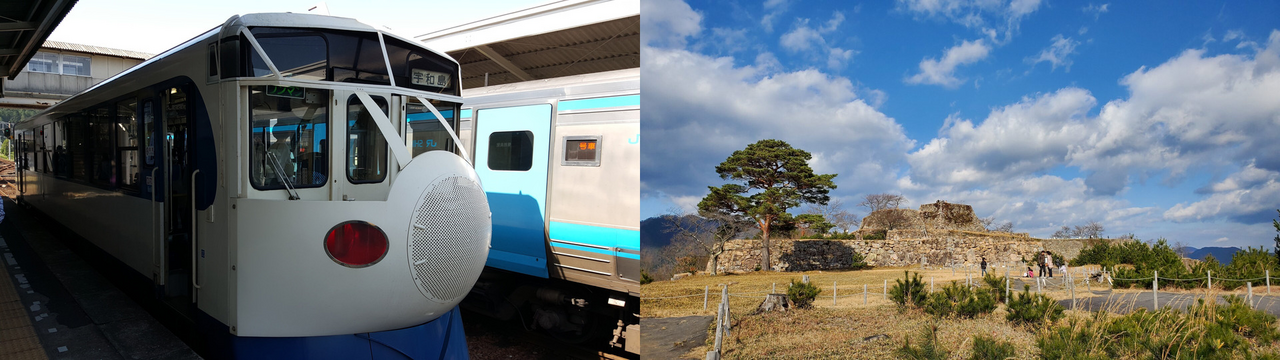}
    \end{minipage}
    &
    \begin{minipage}{0.45\linewidth}
      \includegraphics[width=\linewidth]{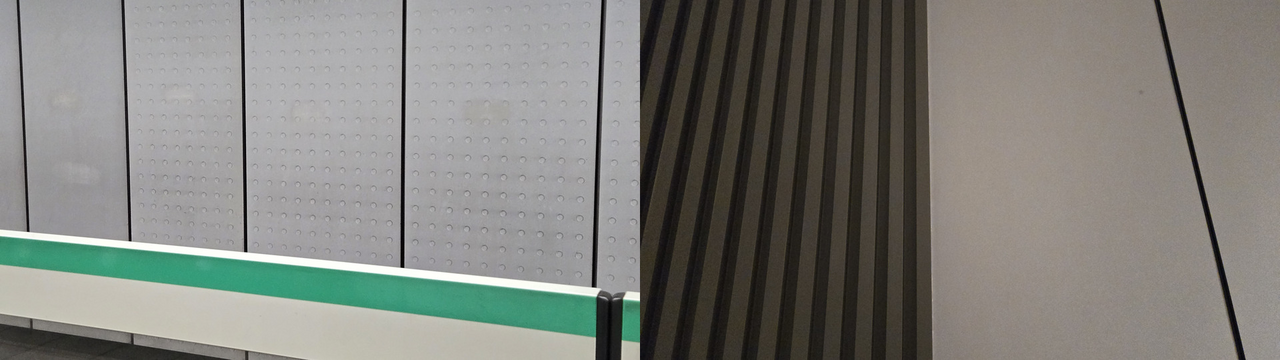}
    \end{minipage}
  \end{tabular}
  \caption{(Left) Examples from the normal dataset, where SIFT is expected to perform well. (Right) Examples from the linear dataset, where strong linear structures make registration challenging for SIFT.}
  \label{fig:datasets}
\end{figure}

We prepared two image sets (normal and linear), each comprising 20 images. \figurename~\ref{fig:datasets} shows examples. The normal dataset contains diverse scenes where SIFT is expected to succeed, whereas the linear dataset contains scenes dominated by strong linear structures where SIFT typically fails. We generated five synthetic pairs per image, yielding 100 pairs per dataset.

Synthetic image pairs and ground-truth homographies were generated as follows. First, a $854\times 480$ pixel image is prepared. A ground-truth homography is then applied such that the four corners are randomly shifted within $p$ pixels, where $p \in \{10,20,30,40,50,60\}$ controls difficulty. Finally, the central $640\times 360$ region is cropped from both images to produce a test pair; the original image is image 1 and the transformed image is image 2.

The geometric error is defined as the average distance between the four corners of image 1 transformed by the estimated homography and by the ground truth. For each perturbation level, we report the mean and standard deviation. Since estimation failure renders this metric uninformative, cases with error exceeding 10 pixels are excluded from the statistics. To evaluate robustness, we also report the success rate, defined as the fraction of successful estimations.

The photometric error is the root-mean-square error (RMSE) between transformed image 1 and image 2 in their overlapping region. Larger photometric error indicates higher risk of visual artifacts such as ghosting. Failed estimations can produce uninformative values; therefore, we introduce a fallback: if the overlap area is less than 100 pixels or if the error exceeds that of the identity transformation (i.e., no warp), we use the identity-transformation value. This choice reflects common commercial practice, where failed homographies are replaced by the identity transform to avoid severe artifacts. Hence, the photometric error better captures practical registration quality.

\begin{figure*}[!htb]
  \centering
  \begin{tabular}{cc}
  (a) &
  \begin{minipage}{0.9\linewidth}
    \includegraphics[width=\linewidth]{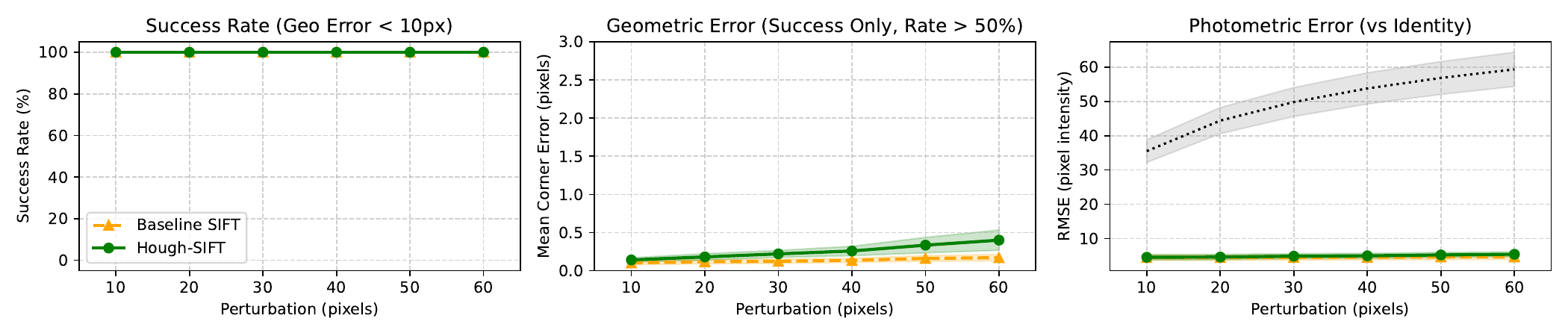}
  \end{minipage} \\
  (b) &
  \begin{minipage}{0.9\linewidth}
    \includegraphics[width=\linewidth]{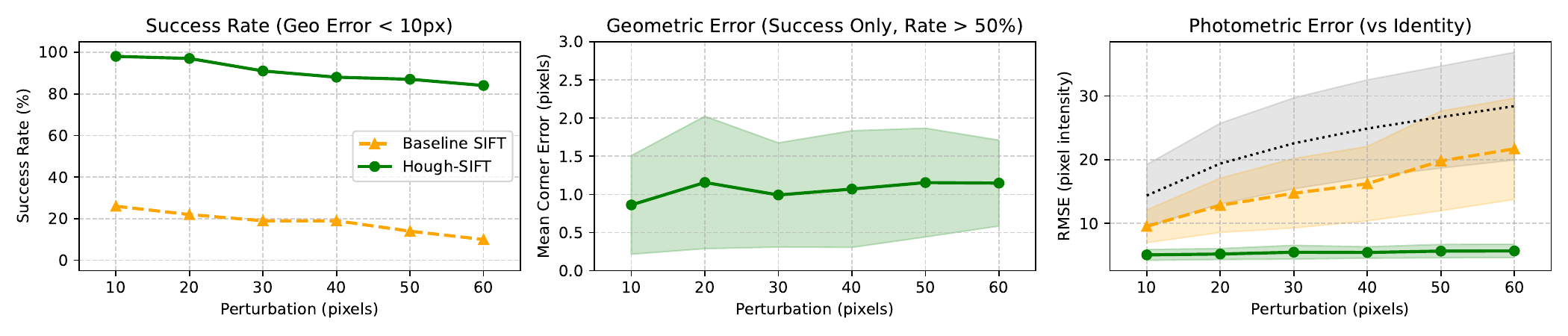}
  \end{minipage}
  \end{tabular}
  \caption{The means and standard deviations of geometric and photometric errors for different values of perturbation $p$. The geometric error is plotted only for data with a success rate greater than 50\%. The black dotted line in the photometric-error graph indicates the error of the identity homography. Each data point was computed from 100 synthetic image pairs. (a) Normal dataset: both SIFT and Hough-SIFT performed well, with SIFT slightly outperforming Hough-SIFT in geometric error. However, Hough-SIFT achieved photometric errors comparable to those of SIFT, indicating similar registration accuracy. (b) Linear dataset: SIFT's performance degraded significantly for all $p$ values, while Hough-SIFT maintained a high success rate and low errors, demonstrating robustness against strong linear structures.}
  \label{fig:graph}
\end{figure*}

\figurename~\ref{fig:graph}(a) shows results for the normal dataset. Both methods performed well, with standard SIFT slightly outperforming Hough-SIFT in geometric error. Because Hough-SIFT performs matching in the line domain, slightly lower point-domain geometric accuracy is expected. However, its photometric error is as low as SIFT's, indicating comparable registration quality in normal scenes.

Results for the linear dataset appear in \figurename~\ref{fig:graph}(b). As expected, SIFT degraded sharply across all $p$ values, with low success rates and large errors. In contrast, Hough-SIFT maintained high success rates and small errors. Notably, it achieved nearly 100\% success for small perturbations ($p=10,~20$), which are typical in applications such as EIS. Even for larger perturbations, Hough-SIFT substantially exceeded SIFT performance, confirming robustness in linear scenes.

Hough-SIFT maintained low photometric errors across all $p$ values, although its success rate degraded at higher $p$. This is likely attributable to the line aperture problem: estimating motion along line direction is inherently ambiguous. The consistently low photometric errors nonetheless indicate reliable registration quality despite this directional ambiguity.

\subsection{Real-World Videos}

\begin{figure}[!htb]
  \centering
  \begin{tabular}{cc}
  \begin{minipage}{0.28\linewidth}
    \includegraphics[width=\linewidth]{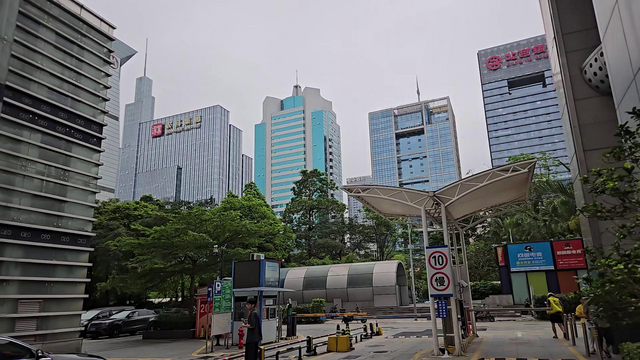}
  \end{minipage} &
  \begin{minipage}{0.28\linewidth}
    \includegraphics[width=\linewidth]{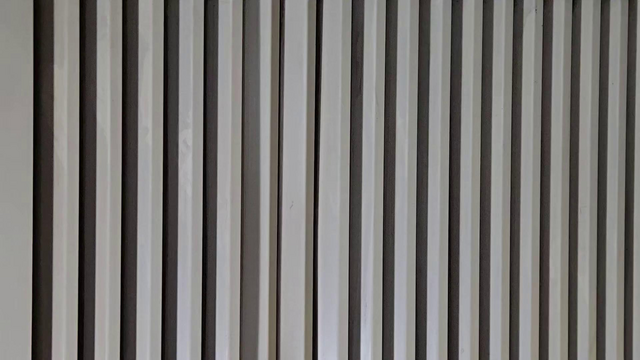}
  \end{minipage}
  \end{tabular}
  \caption{Example frames from the real-world videos used for evaluation. The left panel shows a normal scene, while the right panel shows a scene with strong linear structures.}
  \label{fig:video_frames}
\end{figure}

Next, we evaluated Hough-SIFT on real-world videos of a normal scene and a linear scene. \figurename~\ref{fig:video_frames} shows example frames. Since ground-truth homographies are unavailable for real videos, we used photometric error between consecutive frames as a registration quality proxy. For frame $n$, we estimate the homography $H$ warping frame $n$ to align with frame $n+1$, then compute the photometric error between the warped image and frame $n+1$. If this error exceeds that of the identity transformation or if overlap is below 25\% of the frame, we substitute the identity homography.

\begin{figure}[!htb]
  \centering
  \includegraphics[width=0.85\linewidth]{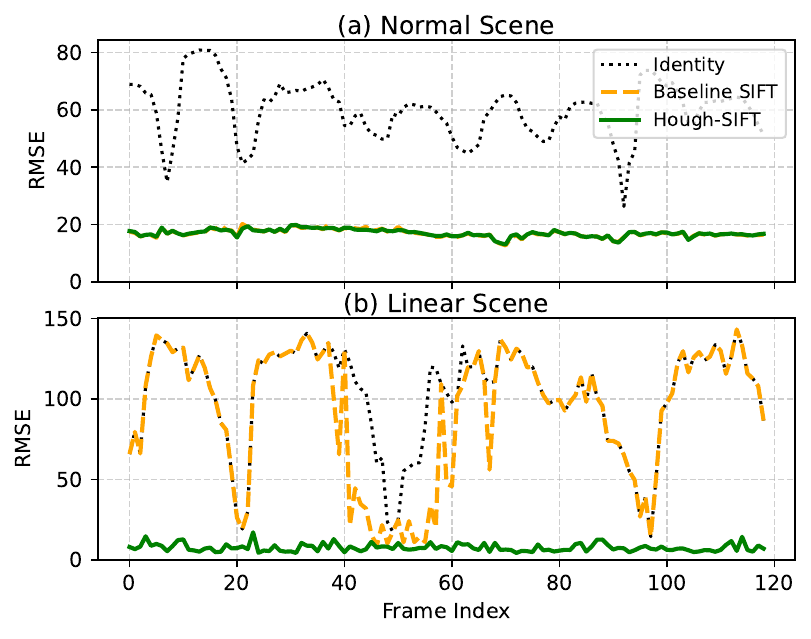}
  \caption{
  Photometric errors for consecutive frame pairs in the real-world videos. SIFT performed well on the normal video but had high errors on the linear video, while Hough-SIFT maintained low errors on both videos, demonstrating its robustness to linear structures in real-world scenarios. Note that the SIFT and Hough-SIFT curves for the normal video almost overlap.
  }
  \label{fig:video}
\end{figure}

\figurename~\ref{fig:video} shows the results. SIFT maintained low errors in the normal scene but degraded in the linear scene, producing large errors. Conversely, Hough-SIFT achieved low photometric errors across both sequences, demonstrating robustness to linear structures in real-world video while preserving performance in normal scenes.\footnote{For visual comparison, please refer to the supplementary video.}

\section{Conclusion}

Hough-SIFT performs SIFT matching in Hough space, enabling robust image registration in scenes with strong linear structures while maintaining accuracy comparable to conventional SIFT in normal scenes. This approach can improve practical computer vision systems. EIS is a representative application: it frequently encounters linear scenes, and registration failures can produce severe visual artifacts, such as unnatural shake and distortion. As smartphone cameras continue to adopt stronger zoom capabilities, this challenge becomes increasingly important. Accordingly, Hough-SIFT can significantly improve robustness in EIS and related applications where linear structures degrade conventional methods.

One limitation of Hough-SIFT is the additional computational cost of the Hough transform. However, this cost can be mitigated by adjusting the Hough-space resolution. The proposed texture-preserving Hough transform is SIMD-friendly and can be efficiently implemented on modern processors.

Future work includes several promising directions. First, developing feature detection and description methods specifically designed for Hough space may yield further performance improvements, since SIFT is optimized for image-space invariance rather than Hough-space properties. Second, a hybrid model combining point-domain and line-domain correspondences in a unified optimization framework could enhance robustness and accuracy. Third, investigating the noise tolerance of the accumulation-based Hough transform in Eq.~(\ref{eq:hough_transform}) may reveal inherent noise suppression properties. Finally, extending the method to three-dimensional applications such as Structure-from-Motion is an interesting direction, although line-based correspondences present fundamental constraints for 3D recovery.

\section*{Acknowledgment}
The author thanks Hiroshi Miyake for an early discussion on matching in Hough space, Ryotaro Kakuda for reviewing the manuscript, Takeshi Miura for fruitful discussions, and Masaki Hilaga for his support and permission to conduct this research as part of corporate duties. An AI language model was utilized to assist with English phrasing and the structural organization of this manuscript.

\bibliographystyle{IEEEtran}
\bibliography{references}

@article{lowe2004sift,
  title={Distinctive image features from scale-invariant keypoints},
  author={Lowe, David G},
  journal={International Journal of Computer Vision},
  volume={60},
  number={2},
  pages={91--110},
  year={2004},
  publisher={Springer}
}

@article{duda1972hough,
  title={Use of the Hough transformation to detect lines and curves in pictures},
  author={Duda, Richard O and Hart, Peter E},
  journal={Communications of the ACM},
  volume={15},
  number={1},
  pages={11--15},
  year={1972},
  publisher={ACM New York, NY, USA}
}

@article{fischler1981ransac,
  title={Random sample consensus: a paradigm for model fitting with applications to image analysis and automated cartography},
  author={Fischler, Martin A and Bolles, Robert C},
  journal={Communications of the ACM},
  volume={24},
  number={6},
  pages={381--395},
  year={1981},
  publisher={ACM New York, NY, USA}
}

@article{canny1986edge,
  title={A computational approach to edge detection},
  author={Canny, John},
  journal={IEEE Transactions on Pattern Analysis and Machine Intelligence},
  number={6},
  pages={679--698},
  year={1986},
  publisher={IEEE}
}

@article{morikawa2021,
    title={Image and video processing on mobile devices: a survey},
    author={Morikawa, Chamin and Kobayashi, Michihiro and Satoh, Masaki and Kuroda, Yasuhiro and Inomata, Teppei and Matsuo, Hitoshi and Miura, Takeshi and Hilaga, Masaki},
    journal={The Visual Computer},
    volume={37},
    pages={2931--2949},
    year={2021},
    publisher={Springer}
}

@book{hartley2003multiple,
    author={Hartley, Richard and Zisserman, Andrew},
    title={Multiple View Geometry in Computer Vision},
    publisher={Cambridge University Press},
    year={2003},
    edition={2nd ed}
}

@article{von2010lsd,
  title={LSD: A line segment detector},
  author={von Gioi, Rafael Grompone and Jakubowicz, Jeremie and Morel, Jean-Michel and Randall, Gregory},
  journal={Image Processing On Line},
  volume={2},
  pages={35--55},
  year={2012}
}

@article{zhang2013lbd,
  title={An efficient and robust line segment matching approach based on LBD descriptor and pairwise geometric consistency},
  author={Zhang, Lilian and Koch, Reinhard},
  journal={Journal of Visual Communication and Image Representation},
  volume={24},
  number={7},
  pages={794--805},
  year={2013},
  publisher={Elsevier}
}

@inproceedings{pautrat2023gluestick,
  title={Gluestick: Robust image matching by sticking points and lines together},
  author={Pautrat, R{\'e}mi and Su{\'a}rez, Iago and Yu, Yifan and Pollefeys, Marc and Larsson, Viktor},
  booktitle={Proceedings of the IEEE/CVF International Conference on Computer Vision},
  pages={9672--9682},
  year={2023}
}

@article{radon1917,
  author  = {Radon, Johann},
  title   = {Über die Bestimmung von Funktionen durch ihre Integralwerte längs gewisser Mannigfaltigkeiten},
  journal = {Berichte über die Verhandlungen der Königlich-Sächsischen Gesellschaft der Wissenschaften zu Leipzig, Mathematisch-Physische Klasse},
  volume   = {69},
  pages    = {262--277},
  year     = {1917}
}

\end{document}